\documentclass[11pt]{article}

\usepackage[final]{acl}

\usepackage{times}
\usepackage{latexsym}

\usepackage[T1]{fontenc}

\usepackage[utf8]{inputenc}
\usepackage{enumitem}
\usepackage{pifont}
\usepackage{amsmath}
\usepackage[utf8]{inputenc} 
\usepackage{CJKutf8}

\usepackage{microtype}

\usepackage{inconsolata}
\usepackage{graphicx}
\usepackage{subcaption}
\usepackage{booktabs}
\usepackage{amssymb}
\usepackage{multirow}   
\usepackage{tabularx}  
\usepackage{xcolor} 

\title{Challenging Multilingual LLMs: A New Taxonomy and Benchmark for Unraveling Hallucination in Translation}

\author{
  Xinwei Wu$^{1,2}$, \quad
  Heng Liu$^2$, \quad
  Jiang Zhou$^{1,2}$, \quad
  Xiaohu Zhao$^2$, \\
  \bf
  Linlong Xu$^2$, \quad
  Longyue Wang$^2$, \quad
  Weihua Luo$^2$, \quad
  Kaifu Zhang$^2$ \\
  \normalfont
  $^1$Tianjin University \quad
  $^2$Alibaba International Digital Commerce \\
}

\begin{document}
\maketitle

\begin{abstract}
Large Language Models (LLMs) have advanced machine translation but remain vulnerable to hallucinations. Unfortunately, existing MT benchmarks are not capable of exposing failures in multilingual LLMs. 
To disclose hallucination in multilingual LLMs, we introduce a diagnostic framework with a taxonomy that separates Instruction Detachment from Source Detachment. 
Guided by this taxonomy, we create \textbf{HalloMTBench}, a multilingual, human-verified benchmark across 11 English-to-X directions. 
We employed 4 frontier LLMs to generate candidates and scrutinize these candidates with an ensemble of LLM judges, and expert validation.
In this way, we curate \textbf{5,435} high-quality instances. 
We have evaluated 17 LLMs on HalloMTBench. Results reveal distinct ``hallucination triggers''---unique failure patterns reflecting model scale, source length sensitivity, linguistic biases, and Reinforcement-Learning (RL) amplified language mixing.
HalloMTBench offers a forward-looking testbed for diagnosing LLM translation failures. HalloMTBench is available in \url{https://huggingface.co/collections/AIDC-AI/marco-mt}.
\end{abstract}

\section{Introduction}
\label{sec:introduction}

Large Language Models (LLMs) have demonstrated remarkable progress in machine translation (MT) \citep{zhao2023survey,chang2024survey}, often surpassing traditional neural machine translation (NMT) systems in fluency and contextual understanding \citep{zhang2023prompting,zhu2023multilingual,zhu2024multilingual}. 
However, their practical deployment remains significantly hampered by the phenomenon of hallucination---where models generate output that is nonsensical or unfaithful to the source text \citep{guerreiro2023hallucinations,huang2025survey,gogoulou2025can}. This issue critically undermines the trustworthiness of LLM-based MT systems and poses a substantial barrier to their real-world application.

\begin{figure}[t!] 
    \centering
    \includegraphics[width=\linewidth]{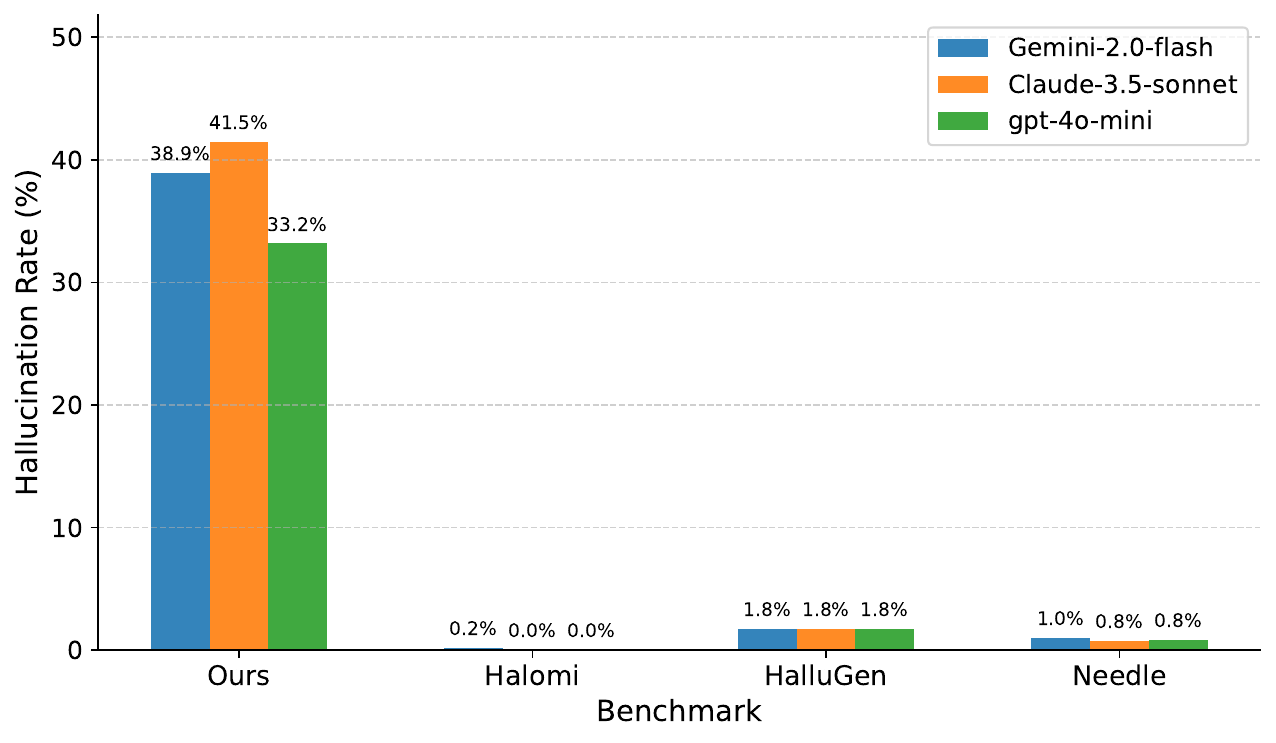} 
    \caption{Obsolescence of existing MT hallucination benchmarks. While leading LLMs achieve a 0\% hallucination rate on established datasets, they exhibit significant hallucination on our proposed benchmark, \textbf{HalloMTBench}. }
    \label{fig:benchmark_obsolescence}
\end{figure}

A critical challenge in addressing LLM hallucinations is the inadequacy of existing evaluation benchmarks \citep{lee2018hallucinations,raunak2021curious,guerreiro2023looking,dale2023halomi}. As illustrated in Figure~\ref{fig:benchmark_obsolescence}, current MT benchmarks are increasingly obsolete, failing to expose the nuanced failure modes of state-of-the-art LLMs. These models can achieve near-zero hallucination rates on traditional evaluations, masking their true vulnerabilities. Compounding this is a lack of a unified and fine-grained definitional framework specifically tailored for LLM-induced translation failures. While previous work has categorized hallucinations in NMT \citep{guerreiro2023hallucinations,zhang2025siren}, the distinct autoregressive and instruction-following nature of LLMs introduces novel and complex hallucination types.

To bridge this gap, we introduce a comprehensive diagnostic framework for LLM translation hallucinations. Our core contribution lies in a refined taxonomy that categorizes these hallucinations into two primary classes: \textbf{Instruction Detachment}, which encompasses failures to adhere to the task instructions (e.g., producing untranslated content or generating output in the wrong target language), and \textbf{Source Detachment}, referring to deviations from the source content (e.g., fabricating extraneous information or generating repetitive text). This taxonomy provides a clear and actionable lens for analyzing LLM translation behaviors.
Guided by this taxonomy, we develop \textbf{HalloMTBench}, a multilingual and human-verified benchmark meticulously designed to stress-test modern LLMs. Spanning 11 English-to-X translation directions, HalloMTBench was curated through a rigorous four-stage process: generating candidate translations using four state-of-the-art LLMs, filtering them with an ensemble LLM Judges approach (which demonstrated $93.68--100$\% agreement with human labels, as detailed in Section \ref{sec:human}), and finally, expert validation. This process ensures the benchmark is both scalable and reliable, resulting in a high-quality dataset of \textbf{5,435} hallucination instances.

Our extensive evaluation of 17 leading LLMs on HalloMTBench reveals distinct ``hallucination triggers''—unique failure patterns that characterize each model's susceptibility to translation hallucinations. These triggers offer critical insights into model-specific weaknesses, uncovering: a consistent performance gap between large proprietary models and smaller open-source MT-LLMs; a U-shaped relationship between hallucination rate and source text length, indicating vulnerabilities at both short and long extremes; and provider- and linguistic biases, such as a home-ground advantage. Critically, reinforcement learning (RL) influenced reasoning appears to amplify language mixing  \citep{guo2025deepseek,wang2025language}, as seen in heightened confusion for RL-tuned models. Thus, HalloMTBench effectively uncovers model-specific vulnerabilities related to model scale, input length sensitivity, linguistic biases, and the impact of training paradigms on reasoning.

In summary, our contributions are threefold:
\begin{itemize} 
    \item We introduce a novel, fine-grained taxonomy for LLM translation hallucinations, precisely distinguishing between Instruction and Source Detachment.
    \item We release \textbf{HalloMTBench}, a large-scale, multilingual, human-verified benchmark specifically designed to challenge and diagnose modern LLMs, along with details on its creation and evaluation methodology.
    \item Through extensive empirical evaluation on HalloMTBench, we uncover and analyze distinctive ``hallucination triggers'', providing critical insights into the reliability of LLM-based translation concerning model scale, training paradigms, source text length, and linguistic biases.
\end{itemize}


\section{Related Work}

\paragraph{Hallucination in NMT}
Early work in the NMT era established a foundational taxonomy of translation hallucinations, broadly categorizing them into two types: hallucinations under perturbation and natural hallucinations \citep{guerreiro2023hallucinations, raunak2021curious}. 
Hallucinations under perturbation occur when minor, artificial disturbances to the source text (e.g., spelling errors) cause the model to generate severely degraded or nonsensical output \citep{lee2018hallucinations}. 
More challenging to detect are natural hallucinations, which appear in translations of clean, unperturbed inputs. 
These are further subdivided into \textit{oscillatory hallucinations}, characterized by the meaningless repetition of words or phrases, and \textit{detached hallucinations}, where the output is fluent but semantically disconnected from the source. 
A subsequent, more fine-grained analysis further distinguishes detached hallucinations into \textit{strongly detached} (partially related to the source) and \textit{fully detached} (completely unrelated) variants \citep{guerreiro2023looking}.

\paragraph{Hallucination in LLMs}
With the advent of Large Language Models (LLMs), the concept of hallucination has broadened to encompass a wider range of failure modes \citep{ji2023survey, huang2025survey}. 
A prominent dichotomy distinguishes between intrinsic and extrinsic hallucinations \citep{gogoulou2025can, zhou2021detecting}. 
Intrinsic hallucinations are attributed to the model's flawed reasoning capabilities, rather than a lack of knowledge. 
Conversely, extrinsic hallucinations stem from factual inaccuracies or gaps in the model's parametric knowledge. 
A related but distinct dichotomy is factuality versus faithfulness \citep{gogoulou2025can}. 
Factuality-based hallucinations refer to discrepancies between the generated content and verifiable real-world facts. 
In contrast, faithfulness-based hallucinations describe divergences from the user's input or a lack of self-consistency within the generated output.

\begin{table*}[ht]
\centering
\caption{Summary of major existing benchmarks for translation hallucination. }
\label{tab:hallucination_benchmarks}
\resizebox{\textwidth}{!}{%
\begin{tabular}{@{}lllll@{}}
\toprule
\textbf{Benchmark} & \textbf{Data Source} & \textbf{Data Size} & \textbf{Generating Models} & \textbf{Language} \\
\midrule
\citet{zhou2021detecting} & \begin{tabular}[c]{@{}l@{}}Patent, COVID-19\end{tabular} & 483  & TranS2S, MBART & Zh, En \\
\midrule
\citet{guerreiro2023looking} & WMT18 & 3,415 & TranS2S & De, En \\
\midrule
\citet{dale2023halomi} & \begin{tabular}[c]{@{}l@{}}FLORES-200\end{tabular} & \begin{tabular}[c]{@{}l@{}}5,247\end{tabular} & NLLB-200-600M & \begin{tabular}[c]{@{}l@{}}9 languages\end{tabular} \\
\midrule
\begin{tabular}[c]{@{}l@{}}\citet{durlich2024overview} \end{tabular} & \begin{tabular}[c]{@{}l@{}}ACES, PAWS-X\end{tabular} & \begin{tabular}[c]{@{}l@{}}400 \end{tabular} & \begin{tabular}[c]{@{}l@{}}Parts of mainstream LLMs \\ (Gemma-7B-it, GPT-3.5/4, \\ Llama-3-it, Llama-2-7b-chat)\end{tabular} & \begin{tabular}[c]{@{}l@{}}En, De, Fr \end{tabular} \\
\midrule
\begin{tabular}[c]{@{}l@{}}\textbf{Ours} \end{tabular} & \begin{tabular}[c]{@{}l@{}}WMT24,HPLT \end{tabular} & \begin{tabular}[c]{@{}l@{}}5,435 \end{tabular} & \begin{tabular}[c]{@{}l@{}}Mainstream LLMs\end{tabular} & \begin{tabular}[c]{@{}l@{}}12 languages\end{tabular} \\
\bottomrule
\end{tabular}%
}
\end{table*}

\paragraph{Benchmarks for Translation Hallucination}
The study of machine translation (MT) hallucinations has led to the creation of specialized benchmarks, which have evolved alongside the models themselves, from traditional NMT systems to modern LLMs.
Early studies were pivotal in establishing foundational definitions and detection methods. For instance, \citet{zhou2021detecting} created a human-annotated dataset for Chinese-to-English translation, while \citet{guerreiro2023looking} developed a detailed hallucination taxonomy for German-to-English using a Transformer-based model.
To broaden linguistic coverage, \citet{dale2023halomi} introduced \textit{HalOmi}, a large-scale, multilingual benchmark generated by the NLLB-200 model.
More recently, the focus has shifted to the unique behaviors of LLMs, with the \textit{HalluciGen} shared task providing a benchmark specifically for models like GPT and Llama \citep{durlich2024overview}.

Table~\ref{tab:hallucination_benchmarks} summarizes these key benchmarks. However, despite their contributions, they are increasingly insufficient for evaluating contemporary models. 
The benchmarks from \citet{zhou2021detecting}, \citet{guerreiro2023looking}, and \citet{dale2023halomi} are based on older NMT architectures. Our preliminary experiments in Fig~\ref{fig:benchmark_obsolescence} reveal that modern LLMs exhibit near-zero hallucination rates on these datasets, rendering them ineffective for differentiating the performance of state-of-the-art models.
Conversely, while the \textit{HalluciGen} dataset targets LLMs, its construction relies on semi-automatic, instruction-based methods. 
This process may not capture the full spectrum of naturally occurring hallucinations.
This gap motivates our work: to create a new, challenging benchmark derived from a diverse set of naturally occurring hallucinations produced by contemporary LLMs, thereby providing a more realistic testbed for future research.

\section{New Taxonomy of Translation Hallucination in LLMs}
\label{sec:Redefining Translation Hallucination}

To establish a precise analytical framework, we first differentiate the hallucinatory patterns exhibited by traditional Neural Machine Translation (NMT) models from those of contemporary Large Language Models (LLMs).

Traditional encoder-decoder NMT models, trained end-to-end for translation, exhibit hallucinations primarily characterized by detachment from the source content.
Conversely, LLMs frame translation as an instruction-following task. Their inherent autoregressive nature, focused on next-token prediction, can override task-specific constraints, giving rise to novel hallucinatory behaviors.
Consequently, prior definitions of hallucination in NMT, which predominantly focused on content fabrication or detachment from the source \cite{lee2018hallucinations, guerreiro2023looking}, are no longer sufficient for the LLM era.
We argue that a comprehensive definition must also encompass the model's failure to adhere to the translation instruction itself.
We therefore propose a more encompassing definition:

\begin{quote}
    \textit{An LLM translation hallucination is any output that deviates from the explicit and implicit constraints imposed by both the translation instruction and the source content.}
\end{quote}

Based on this definition, we introduce a new taxonomy of LLM translation hallucinations, categorizing them into two primary classes: \textbf{Instruction Detachment} and \textbf{Source Detachment}. Table~\ref{tab:hallucination_examples_combined} illustrates this taxonomy with representative examples.

\paragraph{Instruction Detachment} 

This category encompasses hallucinations where the model's output fails to adhere to the explicit or implicit constraints of the translation instruction. We identify two main subtypes:
\begin{itemize}

    \item \textbf{Untranslated Content:} Despite being prompted to translate, the model either repeats the source text verbatim or provides a monolingual paraphrase in the source language.
    \item \textbf{Incorrect Target Language:} The model generates output in a language that is neither the specified target language nor the source language.
\end{itemize}

\paragraph{Source Detachment} 

This category pertains to hallucinations where the generated text, while adhering to instructional constraints (e.g., using the correct target language), contains content unfaithful to the source text. We identify two subtypes:
\begin{itemize}

    \item \textbf{Extraneous Addition:} The model fabricates content that appears as a continuation, elaboration, or speculation on the source text. This is distinct from minor over-translations (or \textit{additions}) in that it introduces semantically significant information completely ungrounded in the source.
    
    \item \textbf{Repetition:} The output contains words or phrases repeated consecutively, often resulting in nonsensical or disfluent text. This is analogous to the \textit{oscillatory hallucinations} observed in earlier NMT models.
\end{itemize}

\begin{table*}[!ht]
\centering
\caption{Examples of the taxonomy for LLM-based translation hallucinations.}
\label{tab:hallucination_examples_combined}
\begin{tabularx}{\textwidth}{p{2.2cm} p{2.2cm} X X}
\toprule
\textbf{Taxonomy} & \textbf{Type} & \textbf{Example} & \textbf{Explanation} \\
\midrule

\multirow{2}{=}{Instruction Detachment} & 
Untranslated & 
\textbf{Source}: me: tbh i liked you blah blah blah \newline 
\textbf{Target}: \textit{Chinese} \newline 
\textbf{Output}: Honestly, I liked you. Blah blah blah. & 
The model was instructed to translate into \textit{Chinese}. Instead, it disregarded the instruction and produced a paraphrase in the source language (English). \\
\cmidrule(l){2-4}
& 
Incorrect Language & 
\textbf{Source}: Small arms (24 September 1999; 26 September 2013). \newline 
\textbf{Target}: \textit{Japanese} \newline 
\textbf{Output}: \begin{CJK*}{UTF8}{gbsn} 小型武器（1999年9月24日；2013年9月26日）\end{CJK*} & 
The instruction specified Japanese as the target language. However, the model generated a translation in \textit{Chinese}, which is neither the source nor the target language. \\

\midrule

\multirow{2}{=}{Source Detachment} & 
Extraneous Addition & 
\textbf{Source}: So I got a Sky Q Mini Box for upstairs. \newline 
\textbf{Target}: \textit{Chinese} \newline 
\textbf{Output}: \begin{CJK*}{UTF8}{gbsn}所以我在天猫上买了Sky Q Mini Box放在楼上。\end{CJK*} & 
The model correctly translated the core sentence but fabricated the detail “(on Tmall)”, which is entirely absent from the source text. \\
\cmidrule(l){2-4}
& 
Repetition & 
\textbf{Source}: Traders Hotel is located within the China World Trade Centre, directly connected to China World Mall and Guomao Metro Station. \newline 
\textbf{Target}: \textit{Chinese} \newline 
\textbf{Output}: 	\begin{CJK*}{UTF8}{gbsn}贸贸贸贸贸贸贸贸贸贸贸贸贸贸贸贸贸贸贸贸贸贸贸贸贸...\end{CJK*}	 & 
After providing a correct translation, the model began to generate nonsensical, repetitive phrases that are detached from the source content. \\

\bottomrule
\end{tabularx}
\end{table*}

\section{Benchmark Curation}
\label{sec:costruc}

To facilitate a systematic study of LLM translation hallucinations, we constructed a large-scale, multilingual benchmark. This benchmark is built upon open-domain translation data and features a meticulous, four-stage annotation process designed to ensure both scale and reliability.

\subsection{Source Corpus}

Our source corpus is derived from the WMT24\footnote{https://www2.statmt.org/wmt24/} shared task materials and HPLT2.0\footnote{https://huggingface.co/datasets/HPLT/}, encompassing 11 high-resource language pairs with English as the source language (EN$\to$\{AR, RU, ZH, JA, ES, FR, DE, PT, IT, KO, VI\}).
For each of these language pairs, we performed random sampling to extract 150,000 sentence pairs from WMT24 and 220,000 from HPLT, thereby creating a massive dataset of 4,070,000 sentence pairs.

\subsection{Annotation Protocol}

Our four-stage annotation framework combines automated generation, ensemble-based detection, and expert validation to achieve both scalability and high annotation reliability.

\paragraph{Stage 1: Large-Scale Translation Generation.}
In this stage, we generated a large-scale corpus of translations using four frontier commercial LLMs: GPT-4o-Mini, Gemini-2.0-Flash, Claude-3.5-Sonnet and Qwen3-Max-20250428.
Each model was tasked with translating the entire 4 million source sentences with temperature set to 0, yielding a total of 16 million translation outputs.
To mitigate potential ordering biases, the execution sequence of the models was randomized for each source sentence.

\paragraph{Stage 2: Ensemble-based Hallucination Detection.}
Subsequently, we employed an automated detection stage inspired by the LLM Judges paradigm. An ensemble of three diverse judge LLMs (including GPT-4o, Claude-3.7-Sonnet, and Gemini-2.5-flash) was deployed to evaluate each generated translation candidate. A translation was identified as a potential hallucination if a supermajority consensus (at least two out of the three judges) was reached on the hallucination label. The specific prompts and detailed criteria employed for this jury voting process are elaborated in the Appendix~\ref{sec:appendix_llm_judge}.

\paragraph{Stage 3: Expert Typology Annotation.}
Following the automated filtering, candidate hallucination instances were meticulously annotated by a team of 5 professional linguists, each possessing extensive experience in MT post-editing and native proficiency in the target languages relevant to the corpus. All annotators underwent a comprehensive 5-hour training program focused on our proposed taxonomy. They utilized a custom-built annotation interface for precise typology tagging and revision tracking. Rigorous ethical research standards were upheld throughout this stage, including competitive compensation for annotators and the procurement of informed consent.

\paragraph{Stage 4: Quality Control.}
A continuous quality control loop ensured annotation reliability. Lead annotators reviewed 10\% of each annotator’s daily submissions, focusing on adherence to taxonomy guidelines and judgment consistency. Inter-annotator agreement (Cohen’s Kappa) was maintained at a minimum of 0.8. Discrepancies were adjudicated by lead annotators, with regular feedback provided to the team. The custom annotation interface facilitated conflict resolution.

\subsection{Benchmark Statistics}
\label{sec:dataset_overview}

Following the four-stage process outlined in Section~\ref{sec:costruc}, our experts validated and categorized the candidate instances, resulting in a final benchmark of \textbf{5,435} high-quality translation hallucinations.

\subsubsection{Data Format}
\label{sec:data_format}

To ensure accessibility and ease of use, all data is provided in a standardized JSON-like format. Each entry in the dataset is a self-contained record that includes the source text, the hallucinated translation, and relevant metadata.

\begin{figure}[htbp]
    \centering
    \includegraphics[width=0.9\linewidth]{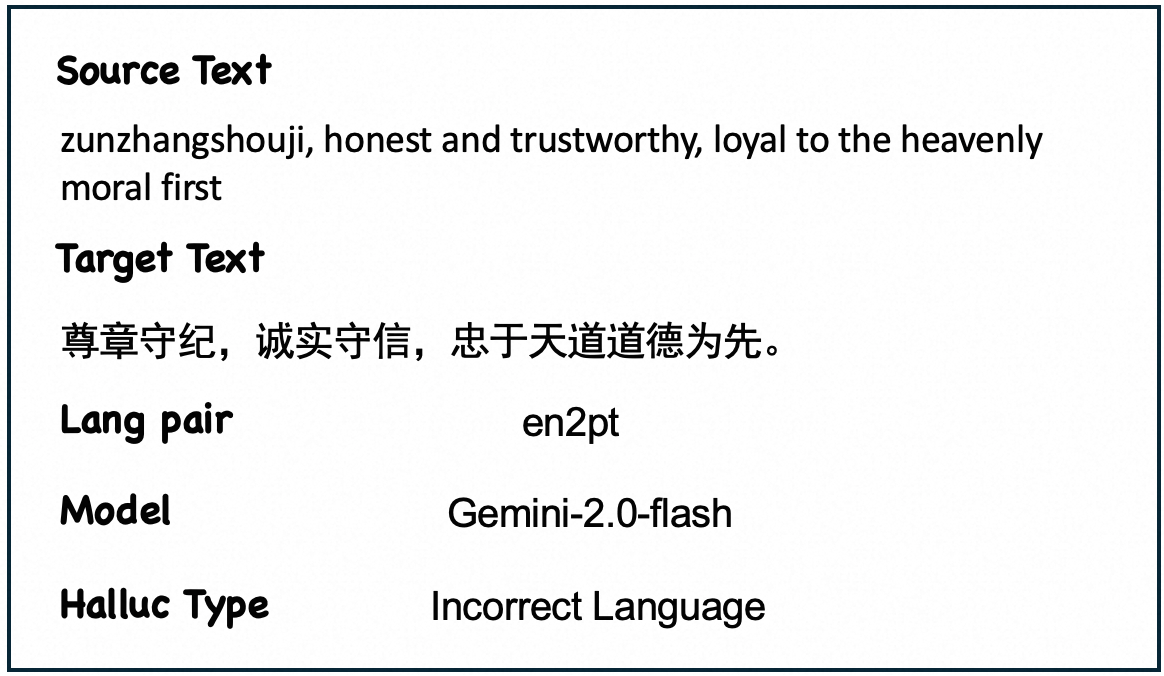} 
    \caption{An example of an ``Incorrect Language'' hallucination data instance from HalloMTBench. }
    \label{fig:data_example}
\end{figure}

Figure~\ref{fig:data_example} illustrates the structure of a single data entry. The key fields are as follows:

\begin{itemize}[leftmargin=*, topsep=3pt, itemsep=0pt]
    \item \textbf{Source Text}: The original text provided to the model for translation.
    \item \textbf{Target Text}: The complete, hallucinated output generated by the model.
    \item \textbf{Language Pair}: The source-to-target language direction (e.g., \texttt{en-pt} for English-to-Portuguese).
    \item \textbf{Model}: The specific LLM that produced the hallucination.
    \item \textbf{Hallucination Type}: The expert-annotated category from our taxonomy (Section~\ref{sec:Redefining Translation Hallucination}).
\end{itemize}

This structured format facilitates straightforward parsing and enables fine-grained analyses of model failure modes.

\subsubsection{Benchmark Composition}

To assess the benchmark's comprehensiveness, we analyzed its composition from two key perspectives: linguistic distribution and hallucination type distribution.

\begin{figure}[t]
    \centering
    \includegraphics[width=0.8\linewidth]{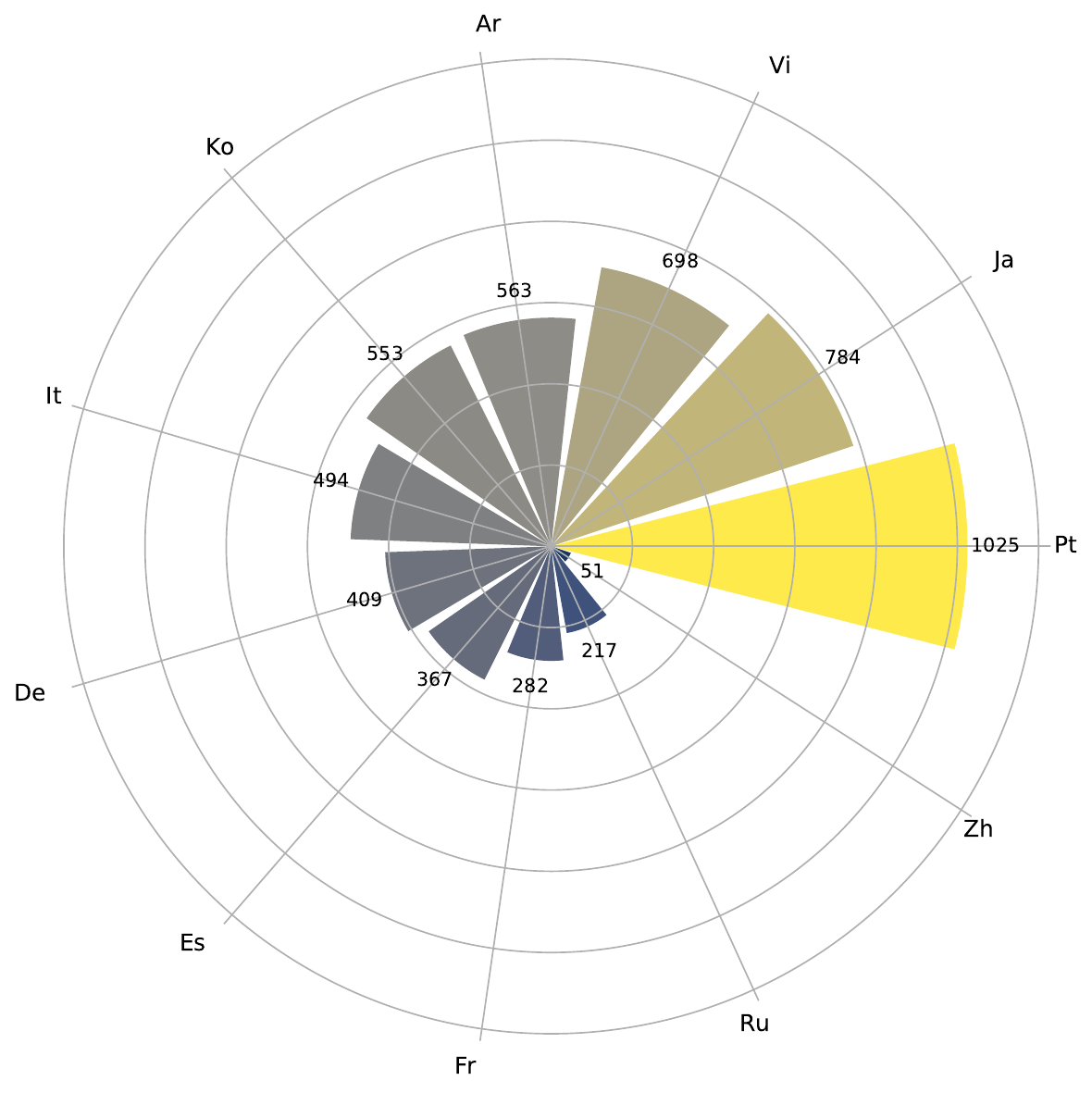}
    \caption{
        Language pair distribution in HalloMTBench dataset. 
        The chart shows the proportion of each English-to-X (`en-xx`) translation direction. 
    }
    \label{fig:lang_distribution}
\end{figure}

\textbf{Linguistic Distribution.} Our benchmark evaluation reveals a stark imbalance in the distribution of hallucination instances across the 11 language pairs, as illustrated in Figure~\ref{fig:lang_distribution}. The incidence of hallucinations is particularly high for Portuguese (\texttt{en-pt}), Japanese (\texttt{en-ja}), and Vietnamese (\texttt{en-vi}), yielding 1,025, 784, and 698 samples, respectively. In stark contrast, the \texttt{en-zh} direction exhibits a remarkably low count of just 51 instances, marking a nearly 20-fold disparity between the highest and lowest counts. This finding highlights a critical, language-specific performance gap and underscores the necessity of broad linguistic coverage in evaluation. Relying on assessments for a few languages, especially those that show low error rates like Chinese in our findings, can mask significant model deficiencies and paint an incomplete, overly optimistic picture of their true capabilities.

\begin{figure}[t]
    \centering
    \includegraphics[width=\linewidth]{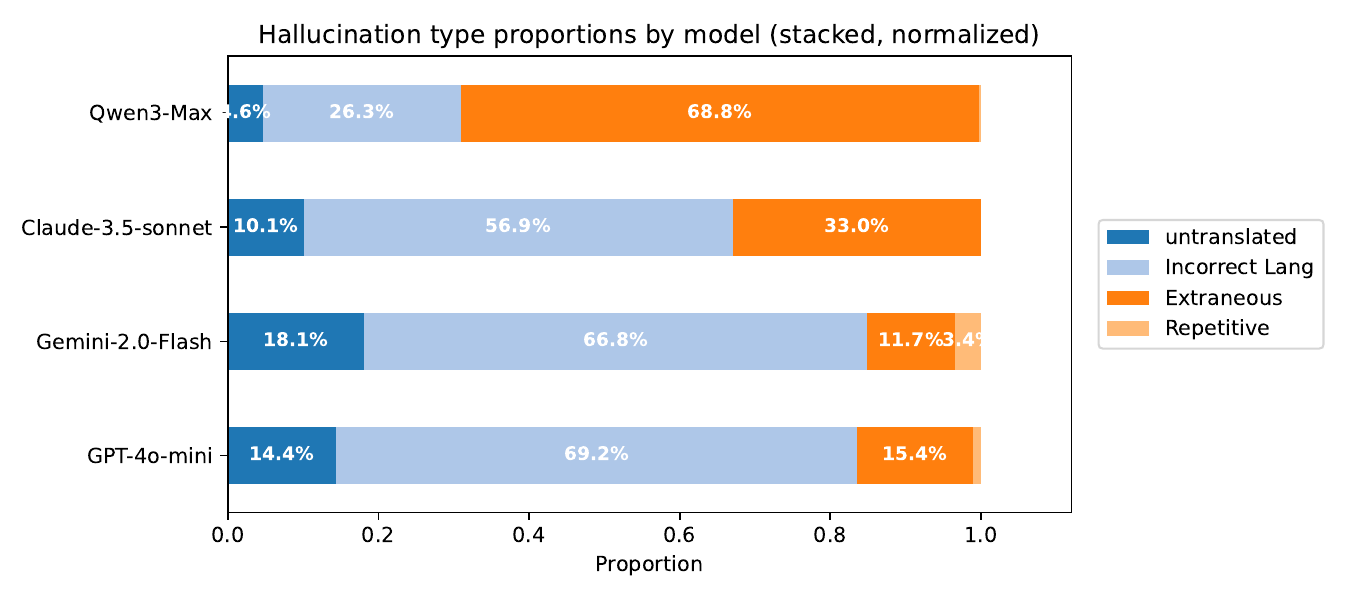}
    \caption{
        Distribution of hallucination types for selected models on our test set. 
        Each stacked bar shows the normalized proportion of different hallucination categories, based on our proposed taxonomy. 
    }
    \label{fig:hallucination_type_distribution1}
\end{figure}

\textbf{Hallucination Type Distribution.} Figure~\ref{fig:hallucination_type_distribution1} presents a fine-grained breakdown of hallucination types by model. While the most prevalent failure modes across all models are generating content in an \textit{Incorrect Language} or producing \textit{Extraneous Additions}, the distribution of these errors is highly model-specific.

For instance, Qwen3-Max exhibits a striking tendency towards \textit{Extraneous Addition}, with this single error type accounting for a remarkable 68.8\% of its hallucinations. This profile is in stark contrast to models like GPT-4o-mini and Gemini-2.0-Flash, which are far more susceptible to generating text in an \textit{Incorrect Language} (69.2\% and 66.8\%, respectively). These distinct "hallucination fingerprints" clearly demonstrate that models fail in fundamentally different ways. Therefore, collecting diverse samples from multiple models is not just a reasonable approach but a necessary one to build a comprehensive and unbiased benchmark.

\section{Experiments and Analysis}
\label{sec:experiments}

To validate the utility of HalloMTBench and diagnose the hallucinatory behaviors of contemporary LLMs, we conducted a large-scale evaluation. This section details our experimental setup and presents a multi-faceted analysis of the results, leveraging our proposed taxonomy to uncover model-specific failure patterns.

\subsection{Experimental Setup}

\paragraph{Models}
We evaluate a diverse set of 17 large language models from five leading developers, representing a broad snapshot of the current LLM landscape. A detailed overview of these models, including their providers and release dates, is provided in Appendix~\ref{tab:llm_overview}.

\paragraph{Translation and Evaluation}
For each of the 17 LLMs, we translated the source texts from HalloMTBench into their respective target languages. All translations were generated under a zero-shot setting with the temperature parameter fixed at 0 to ensure deterministic and high-fidelity outputs. We employed a simple and direct prompt template: ``You are a multilingual expert. Please translate the following text [$source\_text$] from [$source\_lang$] to [$target\_lang$]. Provide only the translation:''.
The maximum number of new tokens was set to 1024.

Based on this LLM-as-judge procedure in Appendix~\ref{sec:appendix_llm_judge}, we calculate our primary metric, the \textbf{Hallucination Rate}, for each model, defined as the percentage of its translations classified into any of the four hallucination types.

\begin{figure}[t!]
    \centering
    \includegraphics[width=\linewidth]{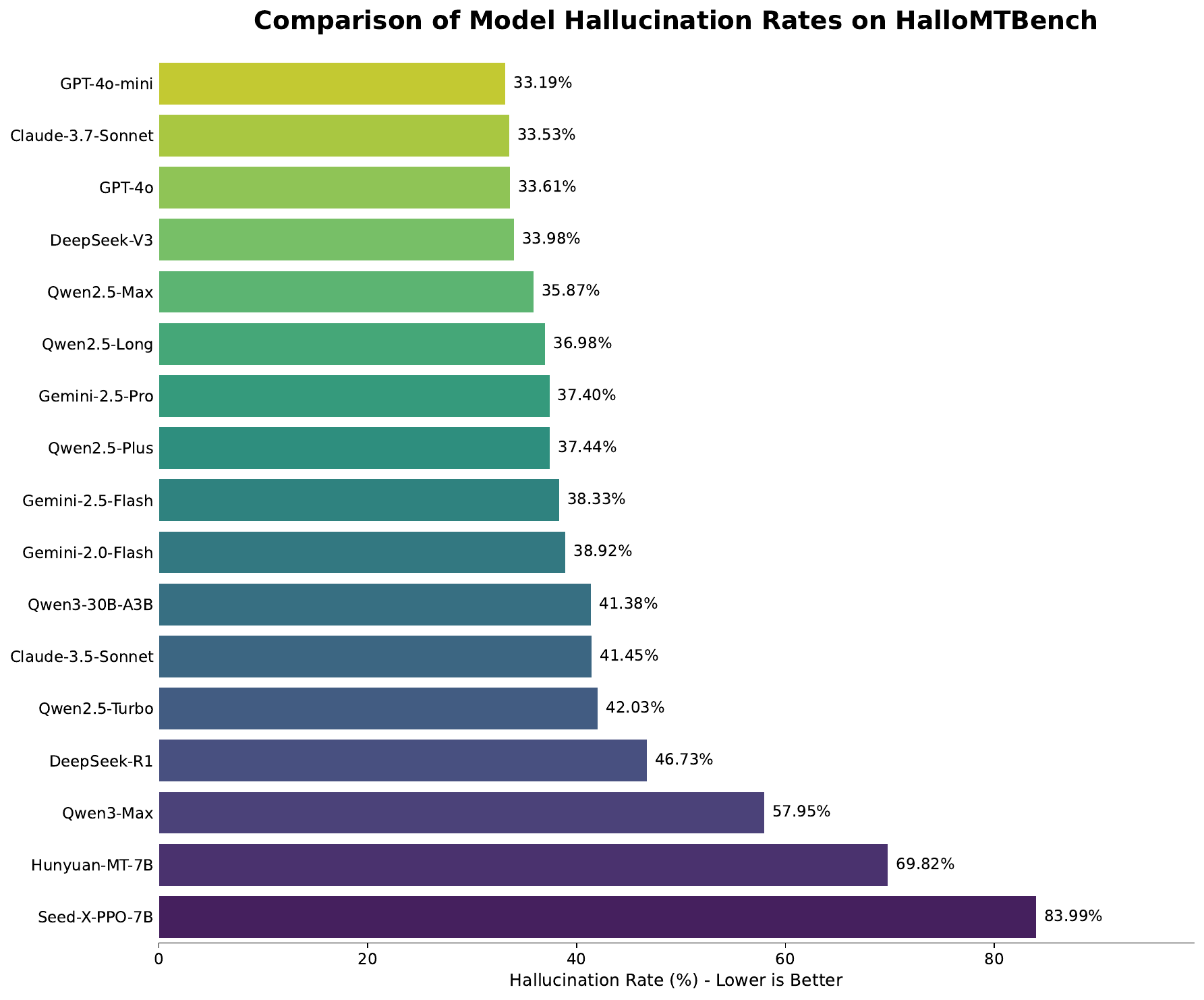}
    \caption{Overall hallucination rates for each model across our entire benchmark.}
    \label{fig:overall_hallucination_rates}
\end{figure}

\subsection{Overall Hallucination Rates}
\label{sec:overall_rates}

Figure~\ref{fig:overall_hallucination_rates} presents the overall hallucination rates for each evaluated model, revealing a wide and significant performance variance across the LLM landscape. The observed rates span from a low of approximately 33\% to a high of nearly 58\%.

Notably, a competitive top-tier of models emerges, exhibiting the lowest tendency to hallucinate. This group is led by GPT-4o-mini (33.19\%), Claude-3.7-Sonnet (33.53\%), and GPT-4o (33.61\%), all clustered closely together in performance. At the other end of the spectrum, Seed-X-PPO-7B shows the highest propensity for hallucination by a substantial margin, with a rate of 83.99\%. This wide spectrum of performance, from the tightly-grouped leaders to the distinct outlier, underscores the benchmark's effectiveness in differentiating model capabilities. It also confirms that susceptibility to translation hallucination remains a pervasive issue, even among otherwise state-of-the-art models.

\begin{figure}[t!]
    \centering
    \includegraphics[width=\linewidth]{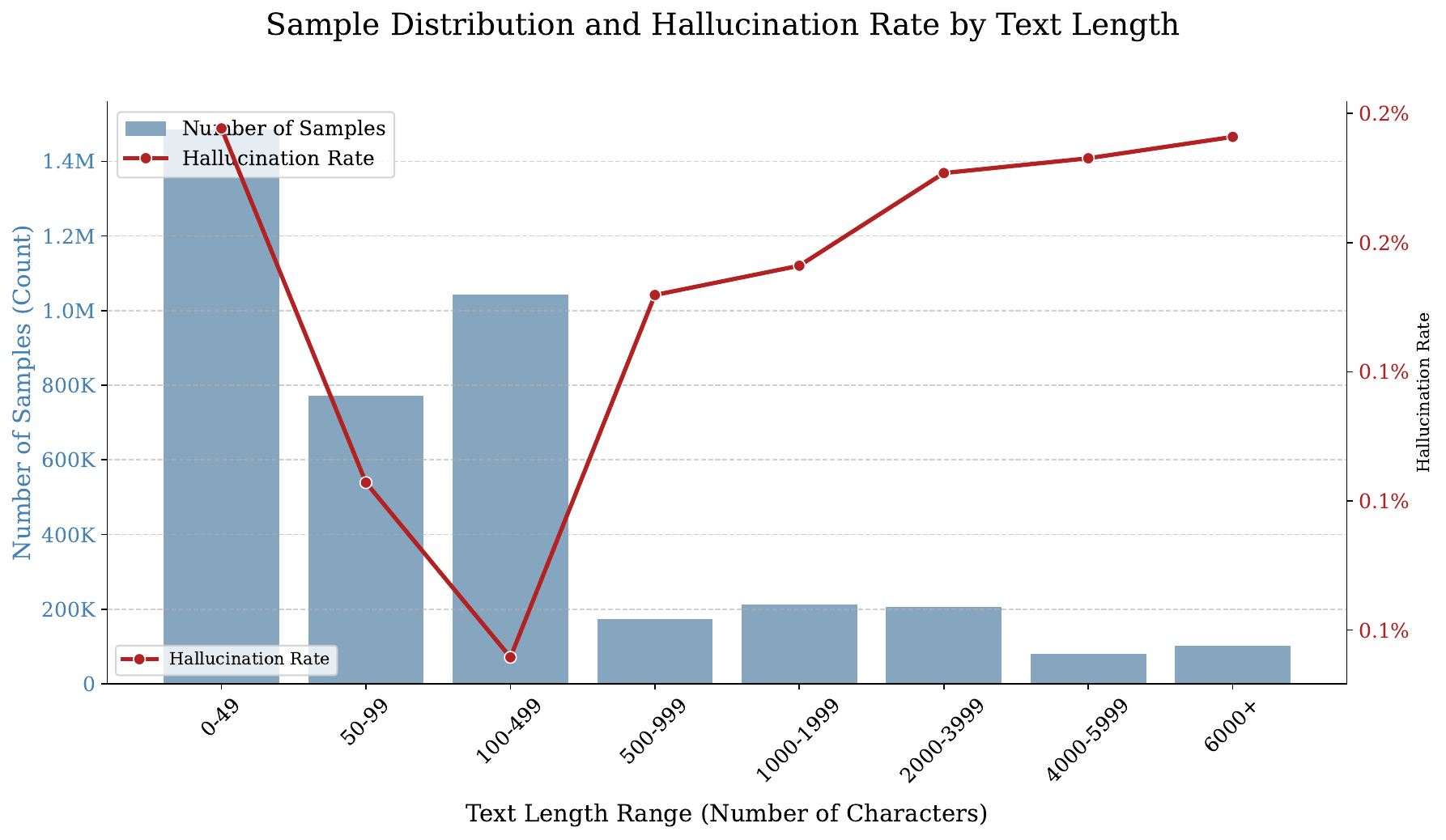}
    \caption{Distribution of Massive Data by source text length (bars) and the corresponding hallucination rate (line).}
    \label{fig:hallucination_length}
\end{figure}

\subsection{Impact of Source Text Length}

To investigate the influence of source text length on translation hallucination, we analyzed the hallucination rate across different length buckets of Massive Data, as illustrated in Figure~\ref{fig:hallucination_length}. Our analysis reveals a distinct U-shaped relationship between the length of the source text and the propensity for hallucination.

The hallucination rate is highest for extremely short texts (0-49 characters), peaking at approximately 0.24\%. As the text length increases, the rate sharply declines, reaching its minimum in the 100-499 character range. Beyond this point, however, a clear trend reversal occurs: the hallucination rate steadily climbs with increasing text length. This suggests that models are vulnerable at both ends of the spectrum. They may struggle with context-deficient short inputs, leading to ungrounded or generic translations, and concurrently fail to maintain factual consistency and long-range dependencies in long-context scenarios, likely due to attentional drift.

\begin{figure*}[t!]
    \centering

    \includegraphics[width=\textwidth]{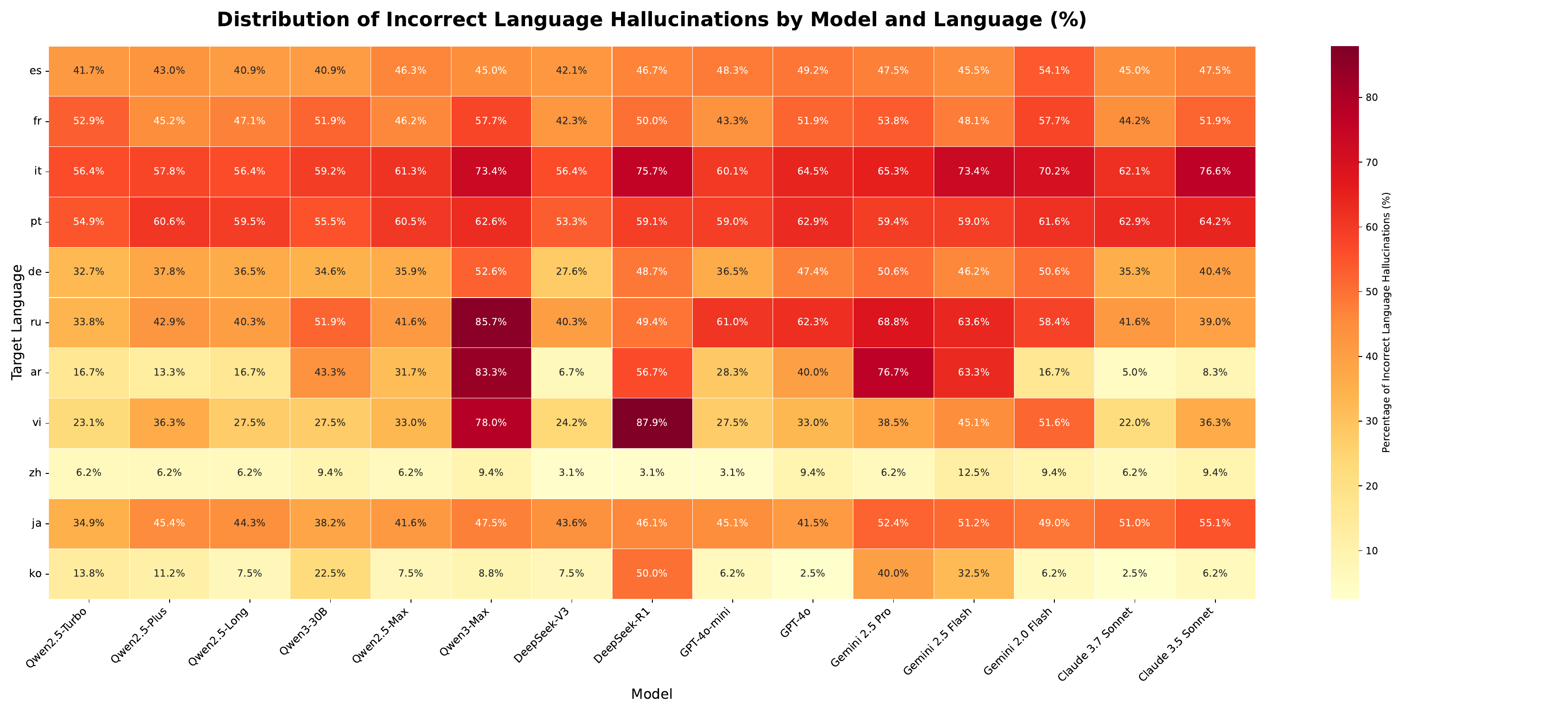}
    \caption{Fine-grained analysis of \textit{Incorrect Language} hallucinations across evaluated LLMs and target languages.}
    \label{fig:benchmark_heatmap}
\end{figure*}

\subsection{Impact of Reasoning RL}
\label{sec:language_biases_rl}

To delve deeper into model failures, we analyzed \textit{Incorrect Language} hallucinations (Figure~\ref{fig:benchmark_heatmap}). 
Comparing base models to RL-aligned counterparts reveals amplified failures: DeepSeek-V3 (base) is confused by \texttt{ar}, while DeepSeek-R1 (RL) has an extreme 87.9\% failure on \texttt{vi}. Similarly, RL-tuned \textbf{Qwen3-Max} shows high concentration on \texttt{ru} (85.7\%) and \texttt{ar} (83.3\%). This aligns with RL-induced \textbf{Language Mixing} \citep{guo2025deepseek}, suggesting RL, while enhancing capabilities, may amplify idiosyncratic failure modes. Evaluating diverse families and paradigms is crucial.

\begin{figure*}[t!]
    \centering
    \begin{subfigure}[b]{0.24\textwidth}
        \centering
        \includegraphics[width=\linewidth]{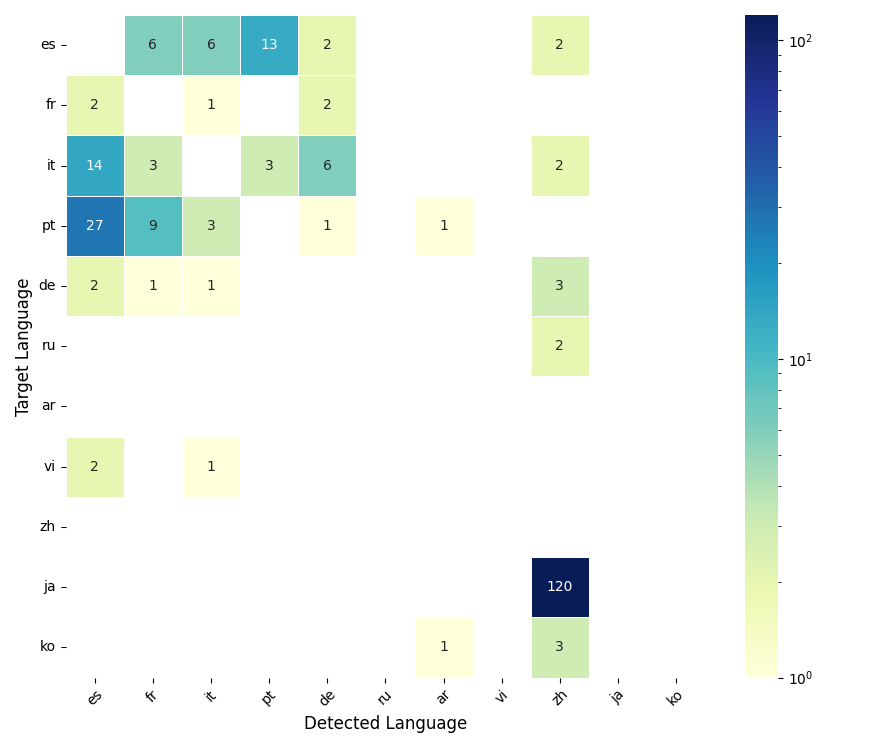}
        \caption{GPT-4o-mini}
        \label{fig:hallucination_gpt}
    \end{subfigure}
    \hfill 
    \begin{subfigure}[b]{0.24\textwidth}
        \centering

        \includegraphics[width=\linewidth]{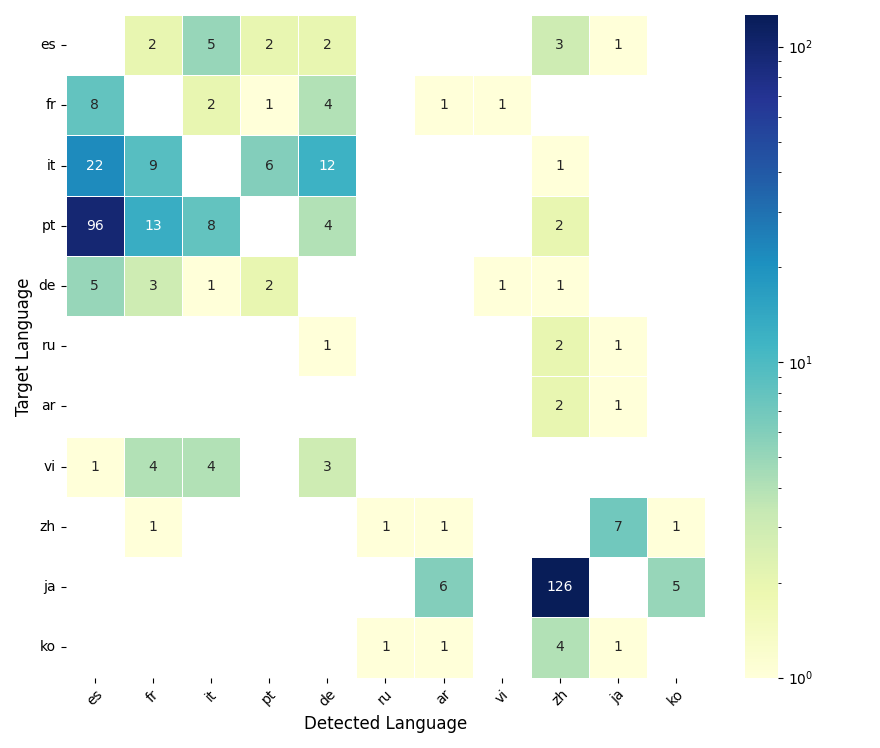}
        \caption{Gemini-2.0-Flash}
        \label{fig:hallucination_gemini}
    \end{subfigure}
    \hfill
    \begin{subfigure}[b]{0.24\textwidth}
        \centering
        \includegraphics[width=\linewidth]{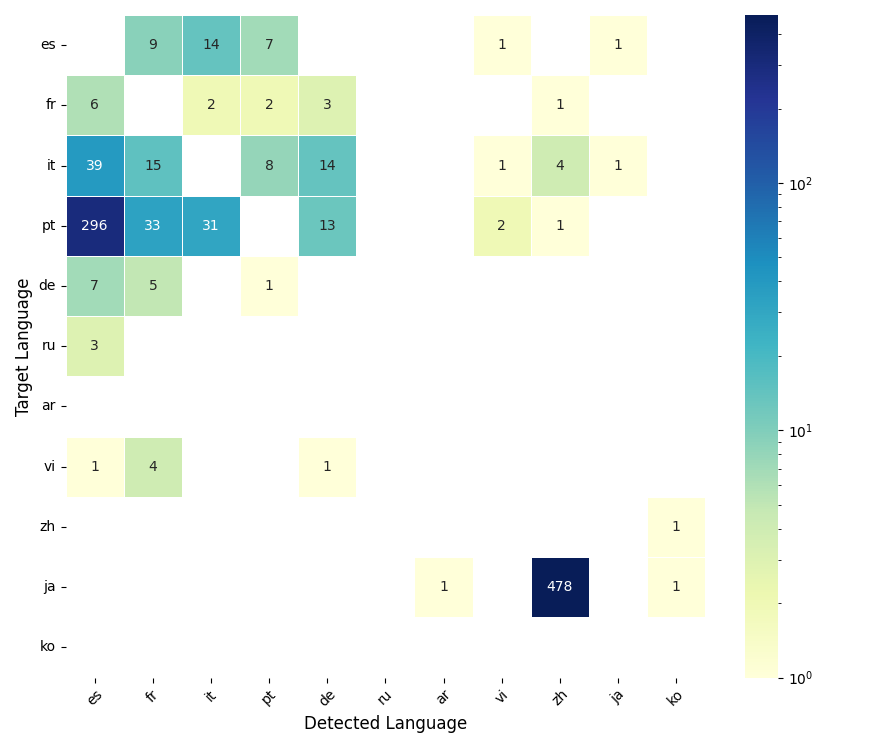}
        \caption{Claude-3.5-Sonnet}
        \label{fig:hallucination_claude}
    \end{subfigure}
    \hfill
    \begin{subfigure}[b]{0.24\textwidth}
        \centering
        \includegraphics[width=\linewidth]{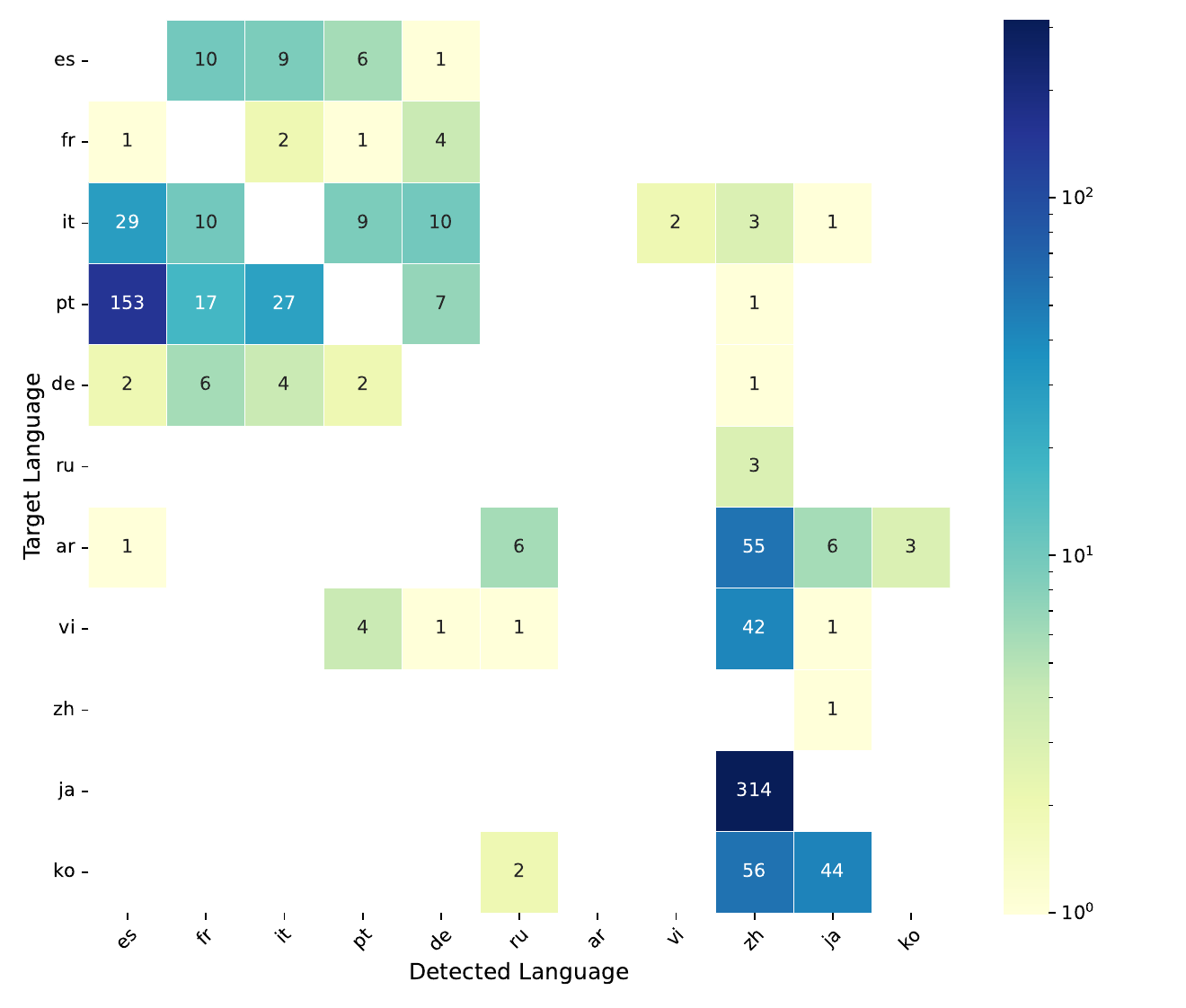}
        \caption{Qwen3-Max}
        \label{fig:hallucination_qwen}
    \end{subfigure}
    \caption{Confusion matrices illustrating language hallucination patterns in English-to-X translation tasks for four proprietary models. The y-axis indicates the intended target language, while the x-axis shows the language of the hallucinated output.}
    \label{fig:hallucination_comparison}
\end{figure*}

\subsection{Impact of Cross-Lingual Bias}
\label{sec:Hallucination Patterns}

To further investigate model vulnerabilities and analyze \textit{Incorrect Language} hallucination patterns, we examined confusion matrices for four leading proprietary LLMs: GPT-4o-mini, Gemini-2.0-Flash, Claude-3.5-Sonnet, and Qwen3-Max (Figure~\ref{fig:hallucination_comparison}). These matrices reveal highly structured, non-random patterns of language confusion that are primarily driven by linguistic similarity, including language family and orthographic overlap.

\textbf{Salient Confusion Clusters Driven by Linguistic Proximity:} Two dominant confusion clusters are consistently observed across these models, directly illustrating the impact of linguistic similarity:

\begin{itemize}
    \item \textbf{Romance Language Confusion:} Models frequently confuse languages with high lexical and syntactic proximity within the Romance family. For instance, Portuguese (\texttt{pt}) targets are often mistranslated or detected as Spanish (\texttt{es}) or other Romance languages. This pattern strongly reflects the deep underlying linguistic relationships and shared vocabulary among these languages.

    \item \textbf{Orthographic Similarity (Chinese/Japanese):} An even more pronounced pattern emerges due to orthographic similarity, particularly the consistent generation of Chinese (\texttt{zh}) for Japanese (\texttt{ja}) targets. This confusion is driven by the shared use of Chinese characters (Kanji). The significantly lower confusion rate with Korean (which uses the distinct Hangul script) further reinforces this hypothesis, highlighting the paramount role of shared scripts in triggering such errors.
\end{itemize}

\textbf{Consistency Across Models and Implications:} Crucially, while the absolute frequencies of these errors vary across models (e.g., Qwen3-Max exhibits notably higher confusion in Romance languages and Japanese targets), the qualitative patterns of confusion remain strikingly consistent. This suggests that the tendency to confuse linguistically similar languages is a fundamental artifact of data-driven representation learning in transformer architectures, rather than an idiosyncrasy of a specific model or its developer. These findings underscore the critical need to consider linguistic typology when evaluating and mitigating translation hallucinations, as these confusions are not random but stem from deep linguistic correlations within the training data.

\begin{figure}[t!]
\centering
\includegraphics[width=0.8\linewidth]{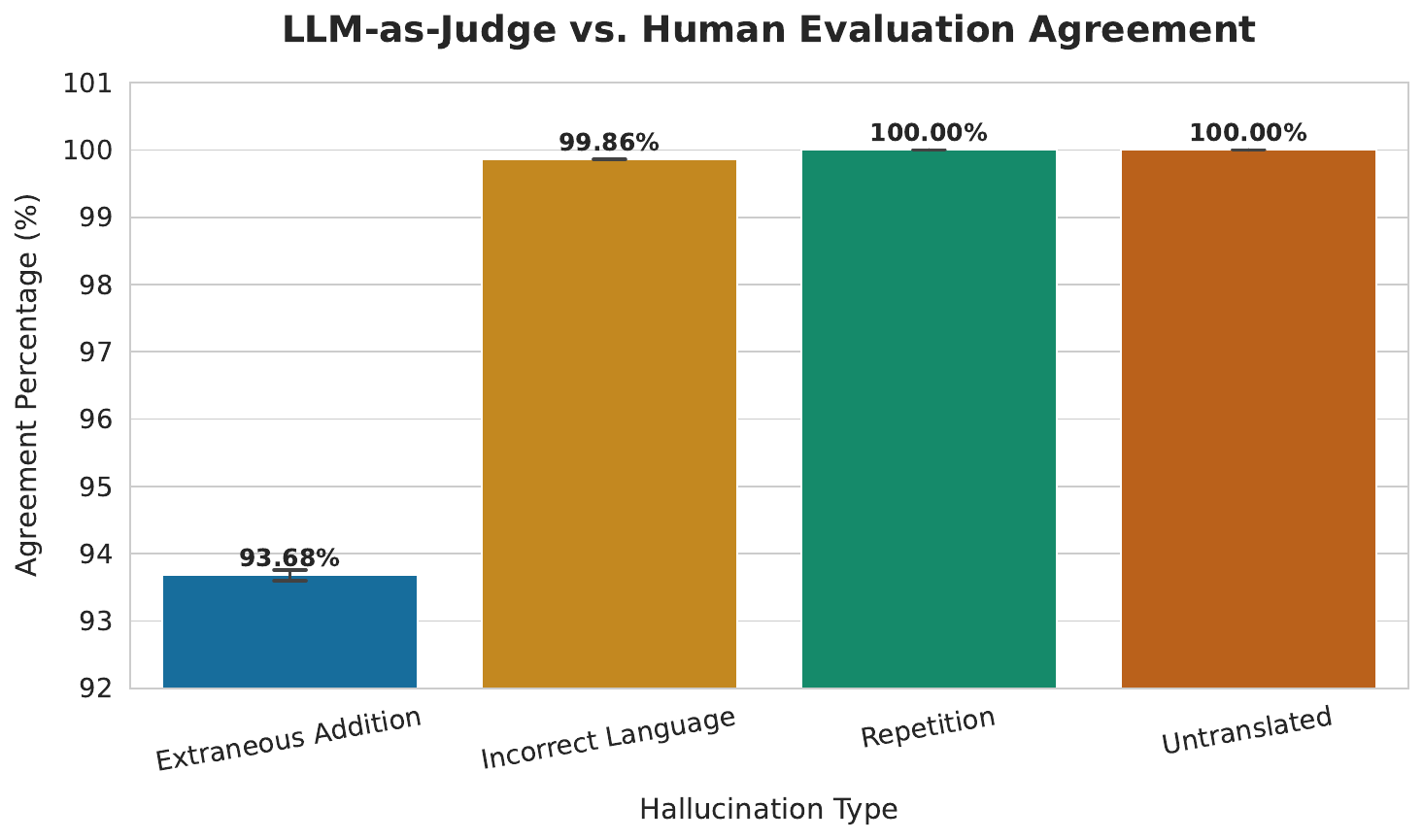}
\caption{Agreement rates between our LLM-as-Judge ensemble and the original human expert labels for each hallucination type.}
\label{fig:llm_human_agreement}
\end{figure}

\subsection{LLM-as-Judge vs. Human Evaluation Consistency}
\label{sec:human}
To rigorously validate our LLM-as-Judge system, we measured its agreement with the original human expert labels for the 5,435 ground-truth samples in \textbf{HalloMTBench}. Our judge ensemble was tasked with re-classifying these samples, and its decisions were directly compared against the human annotations.
As demonstrated in Figure~\ref{fig:llm_human_agreement}, the consistency achieved is exceptionally high. The LLM-as-Judge ensemble achieves near-perfect agreement for objective hallucination types such as \textit{Repetition}, \textit{Untranslated}, and \textit{Incorrect Language} (\textbf{99.86\%–100.00\%}). For the more subjective \textit{Extraneous Addition} category, the agreement remains remarkably strong at \textbf{93.68\%}. This strong correlation validates our LLM-as-Judge as a reliable and scalable proxy for human evaluation, enabling accurate and cost-effective diagnostics without the prohibitive manual effort.

\section{Conclusion}
\label{sec:conclusion}

This paper presents \textbf{HalloMTBench}, a novel, human-verified, multilingual benchmark specifically designed to diagnose hallucination issues in LLM-based machine translation. We introduced a fine-grained taxonomy distinguishing \texttt{Instruction Detachment} from \texttt{Source Detachment}, providing a precise lens for analyzing translation failures.

Our comprehensive evaluation of 17 LLMs on HalloMTBench reveals critical insights into key hallucination triggers. We demonstrate that model scale and type significantly influence overall hallucination rates, with smaller open-source models exhibiting higher susceptibility compared to larger proprietary ones. 
Source text length also acts as a trigger, leading to increased hallucinations in both very short and long inputs. Furthermore, Linguistic biases and RL training induce specific \textit{Incorrect Language} and language mixing patterns, highlighting the impact of training data and paradigms. HalloMTBench effectively uncovers these diverse hallucination triggers and their associated model-specific failure modes, providing a vital testbed for future research aiming to improve LLM translation robustness and trustworthiness.

\bibliography{custom}

\begin{thebibliography}{18}
\providecommand{\natexlab}[1]{#1}

\bibitem[{Chang et~al.(2024)Chang, Wang, Wang, Wu, Yang, Zhu, Chen, Yi, Wang, Wang et~al.}]{chang2024survey}
Yupeng Chang, Xu~Wang, Jindong Wang, Yuan Wu, Linyi Yang, Kaijie Zhu, Hao Chen, Xiaoyuan Yi, Cunxiang Wang, Yidong Wang, and 1 others. 2024.
\newblock A survey on evaluation of large language models.
\newblock \emph{ACM transactions on intelligent systems and technology}, 15(3):1--45.

\bibitem[{Dale et~al.(2023)Dale, Voita, Lam, Hansanti, Ropers, Kalbassi, Gao, Barrault, and Costa-juss{\`a}}]{dale2023halomi}
David Dale, Elena Voita, Janice Lam, Prangthip Hansanti, Christophe Ropers, Elahe Kalbassi, Cynthia Gao, Lo{\"\i}c Barrault, and Marta Costa-juss{\`a}. 2023.
\newblock Halomi: A manually annotated benchmark for multilingual hallucination and omission detection in machine translation.
\newblock In \emph{Proceedings of the 2023 Conference on Empirical Methods in Natural Language Processing}, pages 638--653.

\bibitem[{D{\"u}rlich et~al.(2024)D{\"u}rlich, Gogoulou, Guillou, Nivre, and Zahra}]{durlich2024overview}
Luise D{\"u}rlich, Evangelia Gogoulou, Liane Guillou, Joakim Nivre, and Shorouq Zahra. 2024.
\newblock Overview of the clef-2024 eloquent lab: task 2 on hallucigen.
\newblock In \emph{25th Working Notes of the Conference and Labs of the Evaluation Forum, CLEF 2024. Grenoble. 9 September 2024 through 12 September 2024}, volume 3740, pages 691--702. CEUR-WS.

\bibitem[{Gogoulou et~al.(2025)Gogoulou, Zahra, Guillou, D{\"u}rlich, and Nivre}]{gogoulou2025can}
Evangelia Gogoulou, Shorouq Zahra, Liane Guillou, Luise D{\"u}rlich, and Joakim Nivre. 2025.
\newblock Can llms detect intrinsic hallucinations in paraphrasing and machine translation?
\newblock \emph{arXiv preprint arXiv:2504.20699}.

\bibitem[{Guerreiro et~al.(2023{\natexlab{a}})Guerreiro, Alves, Waldendorf, Haddow, Birch, Colombo, and Martins}]{guerreiro2023hallucinations}
Nuno~M Guerreiro, Duarte~M Alves, Jonas Waldendorf, Barry Haddow, Alexandra Birch, Pierre Colombo, and Andr{\'e}~FT Martins. 2023{\natexlab{a}}.
\newblock Hallucinations in large multilingual translation models.
\newblock \emph{Transactions of the Association for Computational Linguistics}, 11:1500--1517.

\bibitem[{Guerreiro et~al.(2023{\natexlab{b}})Guerreiro, Voita, and Martins}]{guerreiro2023looking}
Nuno~M Guerreiro, Elena Voita, and Andr{\'e}~FT Martins. 2023{\natexlab{b}}.
\newblock Looking for a needle in a haystack: A comprehensive study of hallucinations in neural machine translation.
\newblock In \emph{Proceedings of the 17th Conference of the European Chapter of the Association for Computational Linguistics}, pages 1059--1075.

\bibitem[{Guo et~al.(2025)Guo, Yang, Zhang, Song, Zhang, Xu, Zhu, Ma, Wang, Bi et~al.}]{guo2025deepseek}
Daya Guo, Dejian Yang, Haowei Zhang, Junxiao Song, Ruoyu Zhang, Runxin Xu, Qihao Zhu, Shirong Ma, Peiyi Wang, Xiao Bi, and 1 others. 2025.
\newblock Deepseek-r1: Incentivizing reasoning capability in llms via reinforcement learning.
\newblock \emph{arXiv preprint arXiv:2501.12948}.

\bibitem[{Huang et~al.(2025)Huang, Yu, Ma, Zhong, Feng, Wang, Chen, Peng, Feng, Qin et~al.}]{huang2025survey}
Lei Huang, Weijiang Yu, Weitao Ma, Weihong Zhong, Zhangyin Feng, Haotian Wang, Qianglong Chen, Weihua Peng, Xiaocheng Feng, Bing Qin, and 1 others. 2025.
\newblock A survey on hallucination in large language models: Principles, taxonomy, challenges, and open questions.
\newblock \emph{ACM Transactions on Information Systems}, 43(2):1--55.

\bibitem[{Ji et~al.(2023)Ji, Lee, Frieske, Yu, Su, Xu, Ishii, Bang, Madotto, and Fung}]{ji2023survey}
Ziwei Ji, Nayeon Lee, Rita Frieske, Tiezheng Yu, Dan Su, Yan Xu, Etsuko Ishii, Ye~Jin Bang, Andrea Madotto, and Pascale Fung. 2023.
\newblock Survey of hallucination in natural language generation.
\newblock \emph{ACM computing surveys}, 55(12):1--38.

\bibitem[{Lee et~al.(2018)Lee, Firat, Agarwal, Fannjiang, and Sussillo}]{lee2018hallucinations}
Katherine Lee, Orhan Firat, Ashish Agarwal, Clara Fannjiang, and David Sussillo. 2018.
\newblock Hallucinations in neural machine translation.
\newblock \emph{Transactions of the Association for Computational Linguistics}.

\bibitem[{Raunak et~al.(2021)Raunak, Menezes, and Junczys-Dowmunt}]{raunak2021curious}
Vikas Raunak, Arul Menezes, and Marcin Junczys-Dowmunt. 2021.
\newblock The curious case of hallucinations in neural machine translation.
\newblock In \emph{Proceedings of the 2021 Conference of the North American Chapter of the Association for Computational Linguistics: Human Language Technologies}, pages 1172--1183.

\bibitem[{Wang et~al.(2025)Wang, Lange, Adel, Ma, Str{\"o}tgen, and Sch{\"u}tze}]{wang2025language}
Mingyang Wang, Lukas Lange, Heike Adel, Yunpu Ma, Jannik Str{\"o}tgen, and Hinrich Sch{\"u}tze. 2025.
\newblock Language mixing in reasoning language models: Patterns, impact, and internal causes.
\newblock \emph{arXiv preprint arXiv:2505.14815}.

\bibitem[{Zhang et~al.(2023)Zhang, Haddow, and Birch}]{zhang2023prompting}
Biao Zhang, Barry Haddow, and Alexandra Birch. 2023.
\newblock Prompting large language model for machine translation: A case study.
\newblock In \emph{International Conference on Machine Learning}, pages 41092--41110. PMLR.

\bibitem[{Zhang et~al.(2025)Zhang, Li, Cui, Cai, Liu, Fu, Huang, Zhao, Zhang, Chen et~al.}]{zhang2025siren}
Yue Zhang, Yafu Li, Leyang Cui, Deng Cai, Lemao Liu, Tingchen Fu, Xinting Huang, Enbo Zhao, Yu~Zhang, Yulong Chen, and 1 others. 2025.
\newblock Siren’s song in the ai ocean: A survey on hallucination in large language models.
\newblock \emph{Computational Linguistics}, pages 1--46.

\bibitem[{Zhao et~al.(2023)Zhao, Zhou, Li, Tang, Wang, Hou, Min, Zhang, Zhang, Dong et~al.}]{zhao2023survey}
Wayne~Xin Zhao, Kun Zhou, Junyi Li, Tianyi Tang, Xiaolei Wang, Yupeng Hou, Yingqian Min, Beichen Zhang, Junjie Zhang, Zican Dong, and 1 others. 2023.
\newblock A survey of large language models.
\newblock \emph{arXiv preprint arXiv:2303.18223}, 1(2).

\bibitem[{Zhou et~al.(2021)Zhou, Neubig, Gu, Diab, Guzm{\'a}n, Zettlemoyer, and Ghazvininejad}]{zhou2021detecting}
Chunting Zhou, Graham Neubig, Jiatao Gu, Mona Diab, Francisco Guzm{\'a}n, Luke Zettlemoyer, and Marjan Ghazvininejad. 2021.
\newblock Detecting hallucinated content in conditional neural sequence generation.
\newblock In \emph{Findings of the Association for Computational Linguistics: ACL-IJCNLP 2021}, pages 1393--1404.

\bibitem[{Zhu et~al.(2024)Zhu, Xu, Sun, Pan, Cui, Du, Jin, Branco, Xiong et~al.}]{zhu2024multilingual}
Shaolin Zhu, Shaoyang Xu, Haoran Sun, Leiyu Pan, Menglong Cui, Jiangcun Du, Renren Jin, Ant{\'o}nio Branco, Deyi Xiong, and 1 others. 2024.
\newblock Multilingual large language models: A systematic survey.
\newblock \emph{arXiv preprint arXiv:2411.11072}.

\bibitem[{Zhu et~al.(2023)Zhu, Liu, Dong, Xu, Huang, Kong, Chen, and Li}]{zhu2023multilingual}
Wenhao Zhu, Hongyi Liu, Qingxiu Dong, Jingjing Xu, Shujian Huang, Lingpeng Kong, Jiajun Chen, and Lei Li. 2023.
\newblock Multilingual machine translation with large language models: Empirical results and analysis.
\newblock \emph{arXiv preprint arXiv:2304.04675}.

\end{thebibliography}


\appendix

\section{Appendix}
\label{sec:appendix}

\subsection{The Selection of LLMs}
\label{sec:llm_selection}

To ensure HalloMTBench effectively captures and evaluates translation hallucinations across a wide spectrum of current Large Language Models (LLMs), we adopted a principled selection strategy. 
Our goal is to encompass a diverse set of models that are representative of the state-of-the-art, including both proprietary (closed-source) and open-source architectures of varying scales. 
This diversity is crucial for establishing the broad applicability and long-term relevance of HalloMTBench.

An overview of the selected models, including their developers, parameter counts (where available), release dates in Table~\ref{tab:llm_overview}.
For hallucination generation on WMT data, we focus on four of the most powerful proprietary models: GPT-4o-mini, Gemini-2.0-Flash, Claude 3.5 Sonnet and Qwen3-Max. The rationale for this focused selection is twofold. These models are known for their advanced translation capabilities, meaning their hallucinations are more likely to be subtle and representative of the current frontier of LLM failure cases. Furthermore, as industry-leading models, their outputs define a de facto standard for quality, making their errors critical to benchmark. By using these models to generate translations from WMT source data, we create a challenging testbed for future models to be evaluated against.

\begin{table*}[htbp]
    \centering
    \caption{An overview of the Large Language Models selected for testing translation hallucinations.}
    \label{tab:llm_overview}
    \begin{tabular}{llll}
        \toprule
        \textbf{Company} & \textbf{Model} & \textbf{Parameters} & \textbf{Release Date} \\ 
        \midrule
        \multirow{2}{*}{OpenAI} & \textbf{GPT-4o-mini} & - & 2025-04-14 \\
        & GPT-4o & - & 2024-11-20 \\
        \midrule
        \multirow{2}{*}{Anthropic} & \textbf{Claude-3.5-Sonnet} & - & 2024-10-22 \\
        & Claude-3.7-Sonnet & - & 2025-02-19 \\
        \midrule
        \multirow{3}{*}{Google} & Gemini-2.5-Flash & - & 2025-06-17 \\
        & Gemini-2.5-Pro & - & 2025-03-25 \\
        & \textbf{Gemini-2.0-Flash} & - & 2025-02-05 \\
        \midrule
        \multirow{4}{*}{Alibaba} & Qwen2.5-Plus & - & 2025-04-28 \\
        & Qwen2.5-Long & - & 2025-01-25 \\
        & Qwen2.5-Turbo & - & 2025-04-28 \\
        & Qwen2.5-Max & - & 2025-01-25 \\
        & Qwen3-Max & - & 2025-04-28 \\
        & Qwen3-30B-A3B & 30B & 2025-08-06 \\
        \midrule
        \multirow{2}{*}{Deepseek AI} & DeepSeek-V3 & 671B & 2024-12-25 \\
        & DeepSeek-R1 & 671B & 2025-01-10 \\
        \midrule
        \multirow{1}{*}{Tencent} & Hunyuan-MT & 7B & 2025-09-05 \\
        \midrule
        \multirow{1}{*}{ByteDance} & Seed-X-PPO & 7B & 2025-07-18 \\
        \bottomrule
    \end{tabular}
\end{table*}

\subsection{LLM-as-Judge Evaluation Protocol}
\label{sec:appendix_llm_judge}

This appendix details the LLM-as-Judge protocol employed in this work for evaluating translation hallucinations, complementing the human expert annotations and enabling scalable assessment. Our approach leverages a suite of powerful LLMs to act as expert evaluators, mimicking the decision-making process of human annotators.

\paragraph{Judge Ensemble and Prompts}
We utilize an ensemble of three diverse judge LLMs: GPT-4o, Claude-3.7-Sonnet, and Gemini-2.5-Flash. The rationale for using an ensemble is to mitigate individual model biases and improve the robustness of the evaluation. For each candidate translation, the judges independently assess its quality against our defined hallucination types.

The specific prompts used for each hallucination type are provided in Table \ref{tab:llm_judge_prompts}. These prompts are carefully crafted to elicit binary judgments (presence or absence of the hallucination type) and are parameterized to include the source text, target translation, and relevant language information.

\begin{table*}[t]
\centering
\caption{LLM Judges Prompts for Hallucination Detection}
\label{tab:llm_judge_prompts}
\begin{tabularx}{\textwidth}{|l|X|}
\hline
\textbf{Hallucination Type} & \textbf{Prompt Template} \\
\hline
\texttt{Repetition} & You are a multilingual expert fluent in both [${source\_lang}$] and [${target\_lang}$]. I have a [${source\_lang}$] source text (${source\_text}$) and its [${target\_lang}$] translation [${target\_text}$]. Please check if the translation contains repetitive generation (repeated words or phrases). Return 1 if repetition is present, and 0 if not. \\
\hline
\texttt{Untranslated Content} & You are a multilingual expert fluent in both [${source\_lang}$] and [${target\_lang}$]. I translated a text from [${source\_lang}$] to [${target\_lang}$]. The source text is [${source\_text}$], and the translation is [${target\_text}$]. They appear identical. This can be acceptable for brand names; please judge appropriateness. Return 1 if inappropriate (untranslated), 0 otherwise. \\
\hline
\texttt{Incorrect Target Language} & Please determine the language of the sentence [${target\_text}$] and return its ISO-639-1 abbreviation (en, pl, etc.). If undetermined, output unknown. \\
\hline
\texttt{Extraneous Addition} & You are an expert in detecting translation quality issues. The source text in [\text{${source\_lang}$}] is [$\text{${source\_text}$}$] and the translation in [\text{${target\_lang}$}] is [$\text{${target\_text}$}$]. Your task is to determine if the translation adds specific factual information that is NOT present in the source text. For example, translating 'I got a box' to 'I bought a box on Amazon' is a hallucination. Only return 1 if such fabricated information is found, otherwise return 0. \\
\hline

\end{tabularx}
\end{table*}

\textbf{Decision Rule:} A translation is classified as exhibiting a specific hallucination type if a supermajority (at least 2 out of 3 judges) votes positively (returns 1) for that type. If no type receives a positive vote, the translation is considered faithful. For the \texttt{Incorrect Target Language} type, the predicted language is compared against the expected target language.

\textbf{Consistency with Human Annotations:} As detailed in Section 6.4, our LLM Judges ensemble demonstrated high agreement rates with human expert labels, ranging from 93.68\% for the more subjective \texttt{Extraneous Addition} to 100\% for objective types like \texttt{Repetition} and \texttt{Untranslated Content}. This validates our LLM Judges approach as a reliable and scalable proxy for human evaluation.

\textbf{Open-Source Commitment:} We plan to publicly release the code for our LLM Judges system, including all prompt templates and logic for aggregating judgments. This will enable researchers to leverage this tool for their own evaluations and contribute to the broader understanding of LLM translation quality.

\end{document}